\documentclass[twocolumn]{fairmeta}

\usepackage[table]{xcolor}
\usepackage{adjustbox}
\usepackage{multirow}
\usepackage{tabularx}
\usepackage{makecell}
\usepackage{multirow}
\usepackage{comment}
\usepackage{amsmath}
\usepackage{amsfonts}
\usepackage{xspace}

\newcommand{\method}{{\texttt{WMReward}}\xspace}
\newcommand{\guide}{{\texttt{WMReward($\nabla$+BoN)}}\xspace}
\newcommand{\bonmethod}{\texttt{WMReward(BoN)}\xspace}

\definecolor{darkgreen}{rgb}{0.00, 0.81, 0.78}

\definecolor{rowgray}{gray}{0.95}
\definecolor{SOTAgreen}{RGB}{0,128,0}
\definecolor{SOTAred}{RGB}{180,0,0}
\newcommand{\gain}[1]{\textcolor{SOTAgreen}{\footnotesize\,(+#1)}}
\newcommand{\lose}[1]{\textcolor{SOTAred}{\footnotesize\,(-#1)}}

\newcommand{\eg}{\textit{e.g.}~}

\newcommand{\ie}{\textit{i.e.}~}

\title{Inference-time Physics Alignment of Video Generative Models with Latent World Models}

\author[1,2,*]{Jianhao Yuan}
\author[1,3,4,\ddagger]{Xiaofeng Zhang}
\author[1,\ddagger]{Felix Friedrich}
\author[1,5,\ddagger,*]{Nicolas Beltran-Velez}
\author[1]{Melissa Hall}
\author[1]{Reyhane Askari-Hemmat}
\author[1]{Xiaochuang Han}
\author[1]{Nicolas Ballas}
\author[1,\dagger]{Michal Drozdzal}
\author[1,3,6,7,\dagger]{Adriana Romero-Soriano}

\affiliation[1]{FAIR, Meta Superintelligence Labs}
\affiliation[2]{University of Oxford}
\affiliation[3]{Mila - Qu\'{e}bec AI Institute}
\affiliation[4]{Universit\'{e} de Montr\'{e}al}
\affiliation[5]{Columbia University}
\affiliation[6]{McGill University}
\affiliation[7]{Canada CIFAR AI Chair}

\contribution[*]{Work done at Meta}
\contribution[\dagger]{Joint last author}
\contribution[\ddagger]{Equal contribution}

\abstract{State-of-the-art video generative models produce promising visual content yet often violate basic physics principles, limiting their utility. While some attribute this deficiency to insufficient physics understanding from pre-training, we find that the shortfall in physics plausibility also stems from suboptimal inference strategies. We therefore introduce \method and treat improving physics plausibility of video generation as an inference-time alignment problem. In particular, we leverage the strong physics prior of a latent world model (here, VJEPA-2) as a reward to search and steer multiple candidate denoising trajectories, enabling scaling test-time compute for better generation performance. Empirically, our approach substantially improves physics plausibility across image-conditioned, multiframe-conditioned, and text-conditioned generation settings, with validation from human preference study. Notably, in the \emph{ICCV 2025 Perception Test PhysicsIQ Challenge}, we achieve a final score of $62.64\%$, winning first place and outperforming the previous state of the art by $7.42\%$. Our work demonstrates the viability of using latent world models to improve physics plausibility of video generation, beyond this specific instantiation or parameterization.}

\date{\today}

\metadata[Code]{\url{https://github.com/facebookresearch/WMReward}}
\correspondence{\email{\{adrianars, mdrozdzal\}@meta.com, jianhaoyuan@robots.ox.ac.uk}}

\begin{document}

\maketitle

\section{Introduction}
\label{sec:intro}

\begin{figure*}[t]
  \centering
    \includegraphics[width=\linewidth]{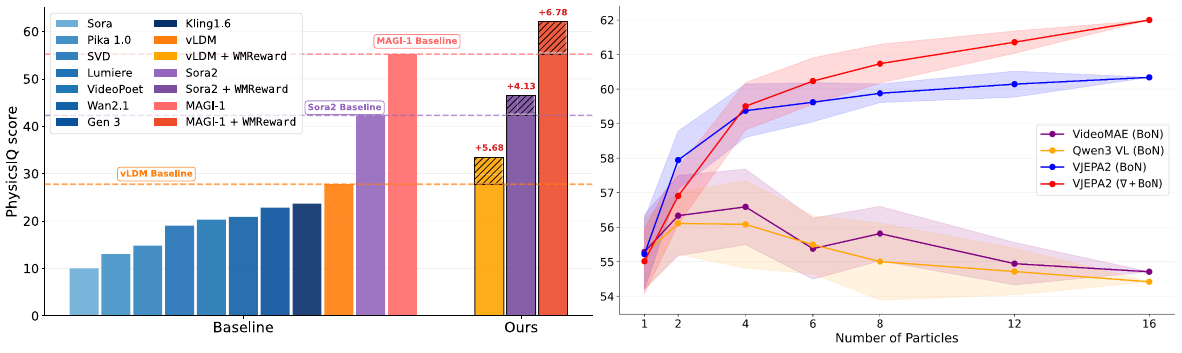}

  \vspace{-3mm}
  \caption{\textbf{Left:}  We achieve the new state-of-the-art on PhysicsIQ benchmark, improving physics plausibility in both single (I2V) and multiframe (V2V) conditioned video generation by a considerable margin. \textbf{Right:} Our latent world model reward \method substantially outperforms VLM and other vision foundation model-based reward signals under BoN search; the proposed \texttt{$\nabla$+BoN} sampling strategy further improves the scaling effect.
  }
  \label{fig:teaser2}
\end{figure*}

State-of-the-art video generative models~\citep{magi1, sora, openai2025sora2,videopoet, lumiere, wan2025, polyak2024movie} have shown remarkable capabilities in generating visually pleasing videos. Yet, the progress in generation quality has been hindered by the limited physics understanding of these models~\citep{kang2024howfar, physicsIQ, yuan2025likephys}, resulting in physically implausible video generation~\citep{bansal2024videophy, bansal2025videophy2}. Ensuring physics correctness in video generation is not only crucial to increase user satisfaction, but also for reliable world modeling~\citep{lecun2022path} and downstream applications such as robotics \citep{yang2023learning} and autonomous driving \citep{hu2023gaia}.\looseness-1

Prior work attributes the observed physical implausibility to the pre-training stage of video generative models~\citep{kang2024howfar}, which relies on minimizing pixel or feature-level reconstruction errors. Following this hypothesis, substantial work has focused on improving pre-training or post-training of video generative models by injecting physics information~\citep{yuan2025newtongen,chefer2025videojam,li2024generativedynamics,cao2024teaching, zhang2025videorepa,zhang2025think}.  By contrast, another line of work assumes that physically plausible videos may be found in the manifold learned by the generative model and therefore has focused on devising inference-time methods to improve physics. This is an underexplored line of research, with two contributions relying on vision-language models (VLMs) to rewrite prompts~\citep{xue2025physt2v} and plan motion~\citep{yang2025vlipp} when a motion-controllable video generative model is available~\citep{burgert2025go}. However, in the image generation literature, alternatives to prompt rewriting~\citep{datta2023prompt,zhang2025intricatedancepromptcomplexity} have been extensively explored to search over the generated data manifold with the goal of boosting performance at inference-time. In particular, reward models have been used to perform candidate selection with search over random seeds~\citep{ma2025inferencescaling, li2025dsearch, zhang2025classicalsearch} and to guide the model's sampling process towards high utility generations \citep{ye2024tfg,hemmat2023feedback,askari2024cvs,dall2025chamfer,askari2025dp,ifriqi2025erg}. 
Despite these advances, existing work has not explored the use of \emph{off-the-shelf} reward models to improve the physics plausibility of video generation at inference time.

Recently, latent world models have demonstrated strong capabilities in physics understanding~\citep{bordes2025intphys,luo2025beyond,garrido2025intuitive}. A \emph{latent world model} is a predictive model that encodes high-dimensional observations (\eg, videos) into compact latent representations and learns the transition function in this latent space to forecast future states. By learning the transition function in this compressed representation rather than in pixel space, these models are trained to focus on features that are predictive of future latent states, such as motion and object dynamics, while ignoring irrelevant appearance details. 
This leads to representations that emphasize fundamental scene properties like structure, object permanence, and trajectory continuity~\citep{garrido2025intuitive}. As a result, they are ideally suited to serve as reward models for physics plausibility. VJEPA-2~\citep{vjepa2} is one particular instantiation of a latent world model, which we employ in this work and which has demonstrated strong physics understanding. Yet, our approach of using latent world models as a source of reward is not tied to this particular instantiation; any suitable latent world model could potentially be used for this purpose.\looseness-1

In this paper, we formulate the problem of improving the physics correctness of video generation as an \emph{inference-time alignment} problem~\citep{uehara2025inference, singhal2025smc} that leverages a \emph{latent world model} with high physics understanding as a reward model to search for physically plausible videos within the manifold learned by the generative model. In particular, we introduce \method by re-purposing VJEPA-2's surprise score as a reward function and show that it can be effectively used both as a Best-of-$N$ (BoN) selector and as a guidance signal during generation to improve the physics of videos. 

We validate the effectiveness of \method by achieving substantial gains in physics plausibility of video generation using three state-of-the-art video generative models: MAGI-1~\citep{magi1}, Sora2~\citep{openai2025sora2}, and a video latent diffusion model (vLDM). These models span diverse architectures, including large-scale autoregressive generation producing videos chunk-by-chunk and holistic diffusion-based generation. We obtain results across text-to-video (T2V), image-and-text-to-video (I2V), and video-and-text-to-video (V2V) settings. In particular, we achieve a new state of the art on the challenging \emph{PhysicsIQ} benchmark~\citep{physicsIQ} with a final score of $62.0\%$, surpassing the previous best by $6.78\%$. This improvement is illustrated by the qualitative examples in \Cref{fig:teaser2} and further validated by an additional $11.4\%$ improvement over baselines in a human-preference study.
Moreover, as shown in \Cref{fig:teaser2}, our VJEPA-based method demonstrates a strong scaling effect with the size of the search space, and outperforms VLM-based selectors (\eg, Qwen3-VL~\citep{yang2025qwen3}) that perform near chance level. With these results, 
we demonstrate the viability of using latent world models to improve the physical plausibility of video generation, paving the way toward developing reliable reward models for video generation inference-time alignment.
\begin{figure*}[ht]
  \centering
    \includegraphics[width=\linewidth]{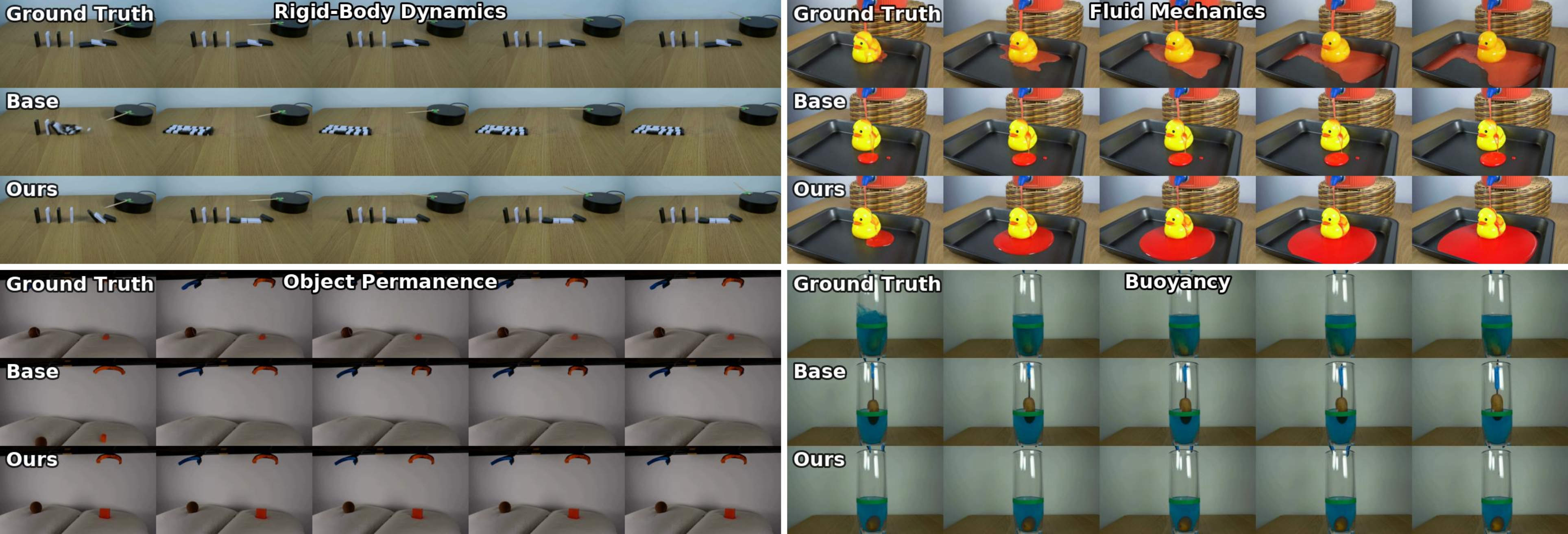}
  \caption{We improve the physics plausibility of video generation by aligning a pre-trained diffusion model with a latent world model at inference time. Using a reward derived from latent world models, we perform search on guided denoising trajectories to sample from a tilted physics plausible distribution. Compared to the baseline (middle row of each quadrant), our generation (bottom) adheres more closely to real-world physics (top), exhibiting smoother temporal continuity, more accurate solid interactions, and improved fluid behavior. 
  }
  \label{fig:teaser1}
\end{figure*}
The key contributions of our work are as follows:\looseness-1
\begin{itemize}
    \item 
    We show that latent world models are useful to improve video generation. To this end, we devise an effective physics plausibility reward model, \method, instantiated with VJEPA's surprise score. 
    \item 
    We highlight the scaling behavior of \method and show that model performance improves when increasing search space -- with guided sampling showing more promising scaling. 
    \item 
    We boost the physics plausibility in video generation across T2V, I2V and V2V settings and datasets, and achieve a final score of 62.0\% on the challenging PhysicsIQ benchmark, outperforming prior state-of-the-art by 6.78\%, which is further validated through a human-preference study showing a boost of 11.4\% in win rate over the baseline. 
\end{itemize}

\section{Methodology: \method}
\label{sec:method}

In the following, we present \method, which transfers the physics prior from a latent world model (\eg, VJEPA-2) into a physics plausibility reward signal. We leverage such reward signal to align the video generative model with the latent world model by sampling from a tilted distribution and thus generate physically more plausible videos.

\subsection{Preliminaries}
Diffusion and flow-matching models are two generative modeling paradigms
that learn to transform Gaussian noise into samples from a target distribution ${p_{\text{data}}(x)}$. They do this by learning score functions $\nabla_{x_t} \log p_t(x_t)$
of the random variables $x_t$ with distribution 
{
\setlength{\abovedisplayskip}{4pt}
\setlength{\belowdisplayskip}{4pt}
\begin{equation}
\label{eq:noise_process}
x_t  = \alpha_t x_0 + \sigma_t \epsilon,
\end{equation}
}\noindent
where $x_0\!\sim\!p_{\text{data}}$ and $\epsilon\!\sim\!\mathcal{N}(0, I)$.  Here,
$\alpha_t$ and $\sigma_t$ are real-valued time-dependent functions where at $t\!=\!0$,
we have $\alpha_0\!=\!1$ and $\sigma_0\!=\!0$ so that $p_0(x)\!=\!p_\text{data}(x)$, 
and at $t\!=\!T$, $p_T(x)$ is approximately Gaussian.
This score is learned using either denoising score matching \cite{vincent2011connection}
or conditional flow matching \cite{lipman2023flow}.

Once this score is available, we can construct an SDE or ODE whose marginals $p_t(x)$ depend only on $\alpha_t$, $\sigma_t$, and the learned score. We then can draw approximate samples from this learned distribution $p(x)$  by solving this SDE/ODE backward from $t=T$ to $t=0$.

\begin{figure}[t]
  \centering
    \includegraphics[width=\linewidth]{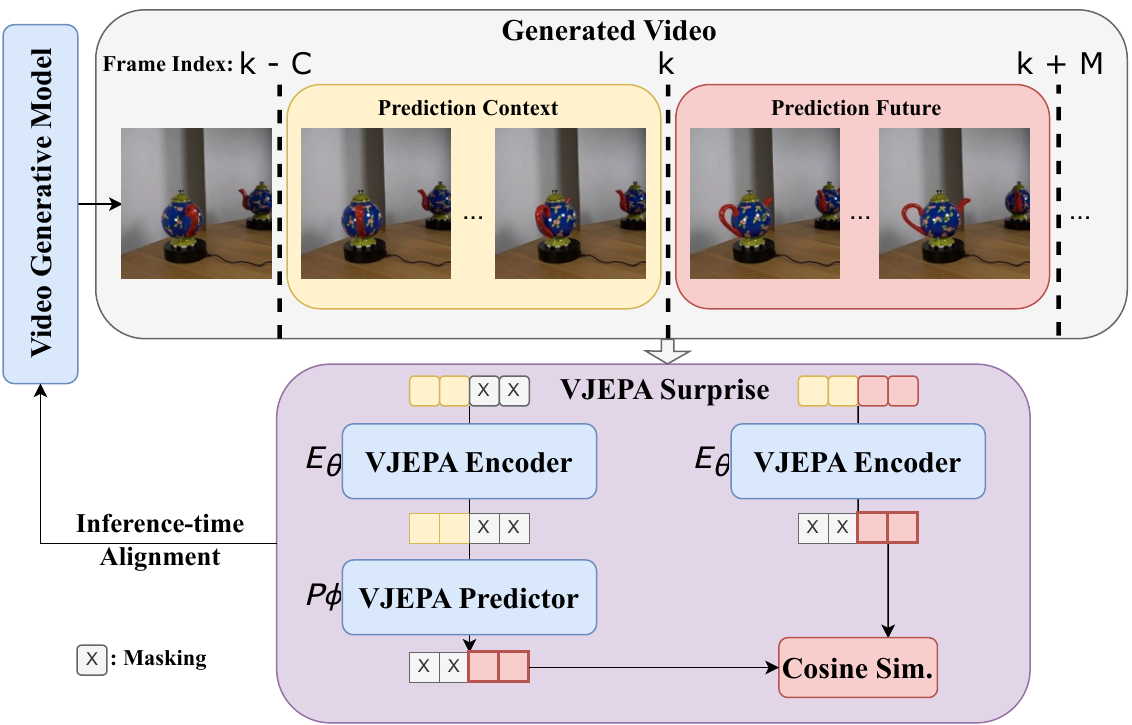}
  \caption{\textbf{Method Overview.} We leverage a latent world model, VJEPA-2, to steer video generative models for better physics plausibility. During generation, we apply a sliding window approach and split the generative model's output into sets of context and future frames. We encode generated context frames and predict the embedding of future frames using the latent world model's predictor. Then, we encode the generated future frames and compute the cosine similarity between its embedding and the latent world model prediction, referred to as surprise score. The surprise score serves as reward to search and guide the denoising trajectories.\looseness-1 
  }
  \label{fig:method}
\end{figure}

\subsection{Steering for Better Physics Plausibility}
\label{sec:steering}
We cast the problem of improving physics plausibility in video generation as 
sampling from a reward-weighted tilted distribution:
\begin{equation}
\label{eq:tilt_dist}
p^*(x) \propto w(x) p(x) ,
\end{equation}
where $p(x)$ is the pre-trained video model distribution and $w(x)\!>\!0$ is a weighting function constructed from the reward $r(x)$ that can evaluate the physics plausibility.

To successfully draw samples from this tilted distribution, two key questions arise: \!1) how to define the reward function that captures physics plausibility from a latent world model (\Cref{sec:vjepa_reward});
and 2) how to realize sampling from $p^*(x)$ given the pre-trained generative model $p(x)$ and the reward function $r(x)$ (\Cref{sec:sampling_schemes}). We depict an overview of our method in \Cref{fig:method}.

\subsection{Latent World Model Reward Signal}
\label{sec:vjepa_reward}

Recent work has demonstrated that latent world models---which learn to make predictions on compressed latent representations rather than raw pixels---can develop the ability to assess physical plausibility \citep{garrido2025intuitive,luo2025beyond,bordes2025intphys}. By operating in a latent space, they tend to ignore superficial visual details and focus on underlying physical dynamics. 
One approach for training such models relies on self-supervised learning; 
 VJEPA-2 \citep{vjepa2} is a latent world model trained in this manner which achieves state-of-the-art performance on physics understanding benchmarks \citep{garrido2025intuitive}, making it an ideal foundation for our reward model.

VJEPA learns representations through self-supervised prediction of unmasked video regions from masked ones. The architecture comprises a context encoder $E_\theta$ that embeds video frames, and a predictor network $P_\phi$ that reconstructs target representations from partial observations. During training, a masked view $x_{\text{masked}}$ is created from a video $x$ by removing certain spatiotemporal regions. The predictor $P_\phi$ takes the encoder's output $E_\theta(x_{\text{masked}})$ along with learnable mask tokens $\Delta_m$, indicating masked positions, and attempts to reconstruct the target representation. The target is computed using an exponentially-moving-averaged (EMA) encoder $\bar{E}_\theta$ applied to the full video. The training objective is:
\begin{equation}
\mathcal{L} = \|P_\phi(\Delta_m, E_\theta(x_{\text{masked}})) - \text{sg}(\bar{E}_\theta(x))\|_1
\end{equation}
where $\text{sg}(\cdot)$ denotes the stop-gradient operation, preventing gradients from flowing through the target. The idea is that by making predictions in feature space, the model is encouraged to learn high-level spatiotemporal features that capture predictable dynamics rather than pixel-level details.

To turn VJEPA into a physics plausibility reward model, we leverage a simple intuition: being a world model, VJEPA should resonably predict the future in physically plausible videos. Hence, the more its predictions diverge from the generations, the more likely it is that the videos are less physically plausible.

We materialize this intuition by designing a reward function, \method, that measures prediction surprise by contrasting VJEPA-2 predictions with the generated video as shown in \Cref{fig:method}.
Concretely, we slide a window of length $C+M$ across a generated video $x$, where $C$ denotes context frames and $M$ denotes prediction horizon. At each position $k$, VJEPA observes context frames $x^{k-C+1:k}$ and must predict representations for future frames $x^{k+1:k+M}$. 
We produce latent world model's future representations using only context, as follows: \looseness-1
{
\setlength{\abovedisplayskip}{4pt}
\setlength{\belowdisplayskip}{4pt}
\begin{equation}
\hat{z}_k = P_\phi(\Delta_m, E_\theta(x^{k-C+1:k})),
\end{equation}
}\noindent
where $\Delta_m$ indicates that $M$ future positions are masked. We obtain generated video's representations by processing the complete window:
{
\setlength{\abovedisplayskip}{4pt}
\setlength{\belowdisplayskip}{4pt}
\begin{equation}
z_k = E_\theta(x^{k-C+1:k+M}).
\end{equation}
}\noindent
Both representations contain representations for all $C+M$ positions; we extract and compare only the future portions corresponding to positions $C+1$ through $C+M$ ($\hat{z}_k^{\text{fut}}$ and $z_k^{\text{fut}}$):\looseness-1
{
\setlength{\abovedisplayskip}{4pt}
\setlength{\belowdisplayskip}{4pt}
\begin{equation}\label{eq:surprise}
r(x) = \frac{1}{|\mathcal{K}|} \sum_{k \in \mathcal{K}} \left(1 - \cos(\hat{z}_k^{\text{fut}}, z_k^{\text{fut}})\right),
\end{equation}
}\noindent
where $\mathcal{K}$ contains all valid window positions. Videos receive higher rewards when their generated futures closely match VJEPA's predictions, effectively measuring physical coherence.

$\mathcal{L} = 1 - \cos(z_{pred}, \;z_{gt})$

\subsection{Instantiation of Sampling Schemes}
\label{sec:sampling_schemes}
We now consider practical sampling schemes with different weighting functions $w(x)$ to draw samples from the defined tilted distribution in \Cref{eq:tilt_dist}.
As presented in \Cref{sec:vjepa_reward}, the reward function $r(x)$ is differentiable with respect to the video $x$, which allows us to leverage both gradient-based and gradient-free sampling schemes. We focus on three sampling schemes chosen for their representativeness, simplicity, and performance: guidance as an example for gradient-based method, best-of-$N$ for gradient-free method, and their combination.

(i) \textbf{Guidance ($\nabla$)} uses 
${w(x) = \exp(\lambda r(x))}$, where $r(x)$ is our reward function and $\lambda > 0$ is a temperature parameter controlling how much we upweight high-reward samples.
Under the same noising process as in \Cref{eq:noise_process},
one can show that the score functions $ \nabla_{x_t} \log p_t^*(x_t)$ of the time-dependent marginals $p_t^*(x_t)$ are given by
\begin{equation}\label{eq:score}
\nabla_{x_t} \log p_t(x_t) + \nabla_{x_t} \log \mathbb{E}\left[ e^{\lambda r(x_0)} ~ \middle| ~x_t \right].
\end{equation}
Therefore, approximating $\mathbb{E}[e^{\lambda r(x_0)} \mid  x_t]$ as $e^{\lambda r(\mathbb{E}[x_0 \mid x_t])}$, and using Tweedie's formula
\citep{robbins1992empirical, efron2011tweedie},

\begin{equation}
x_{0\mid t} := \mathbb{E}\left[
x_0 \mid x_t
\right]
= 
\frac{1}{\alpha_t}
\left( 
x_t + \sigma_t \nabla_{x_t} \log p_t(x_t)
\right),
\end{equation}
we obtain the following approximation to \Cref{eq:score}
\begin{equation}
\label{eq:score-guidance-approx}
\nabla_{x_t}\!\log p_t^*(x_t) \approx \nabla_{x_t}\! \log p_t(x_t) + \lambda \nabla_{x_t}\! r(x_{0|t}(x_t))
\end{equation}
We can therefore sample from $p^*(x)$ by using this new score approximation in the SDE/ODE sampler.

(ii) \textbf{Best-of-$N$ Search} (BoN) involves generating $N$ independent samples $\{x^{(i)}\}_{i=1}^N$ (\ie, particles) from the base model $p(x)$
and selecting the sample with highest reward:
\begin{equation}
 x^* = \text{argmax}_{x \in \{x^{(i)}\}_{i=1}^N} r(x) .
\end{equation}
This procedure effectively samples from a distribution
\begin{equation}
p^*(x) \propto p(x) [F(r(x))]^{N-1} ,
\end{equation}
where $F(\cdot)$ is the cumulative distribution function (CDF) of $r(x)$ under $p(x)$ and $w(x) = [F(r(x))]^{N-1}$.

(iii) \textbf{$\nabla$ + BoN} combines both approaches by using the guidance scheme to generate $N$ samples
and then chooses the highest reward sample among them. 
This yields approximate samples from the tilted distribution with weight function $w(x) = \exp(\lambda r(x)) \cdot [F_\lambda(r(x))]^{N-1}$,
where $F_\lambda(\cdot)$ is the CDF of $r(x)$ under the guided distribution. 
This combination is beneficial because it achieves stronger tilting without requiring large $\lambda$
values (where the score approximation becomes inaccurate), and best-of-$N$ selection  provides an additional strategy for filtering samples where the approximation might have been sub-optimal. In this way, we benefit from the increased likelihood of high-reward samples provided by guidance, while mitigating approximation errors. 
\looseness -1

\section{Experiments}
\label{sec:experiments}

We show the effectiveness of sampling with \method for improving physics plausibility across three video generation setups: image-and-text-to-video (I2V), video-and-text-to-video (V2V), and text-to-video (T2V). For all experiments, we apply \method to MAGI-1 24B~\citep{magi1} and to vLDM 5B\footnote{vLDM is a latent video diffusion model that utilizes a DiT backbone and jointly models spatial and temporal components at once.\looseness-1}, and additionally to Sora2~\citep{openai2025sora2} on the PhysicsIQ benchmark.\looseness-1

\subsection{Image and Multiframe-conditioned Generation}
\label{sec:physicsiq}

We start by evaluating \method on PhysicsIQ \citep{physicsIQ}, a benchmark for I2V and V2V generation. We follow their setup and use the text prompt along with a three-second context video (for V2V) or the last frame (for I2V). Given the conditioning, we generate five-second video continuations, evaluate it against the ground-truth and compute the PhysicsIQ score that relies on an aggregation of four metrics: spatial intersection over union (IoU), spatio-temporal IoU over motion masks, a weighted spatial IoU, and pixel MSE. For all methods, we use 16 particles for the search.\looseness-1

\begin{table}[t]
\centering
\caption{\textbf{Image and Multiframe-conditioned Generation Performance on PhysicsIQ.} 
Our results are highlighted in gray and performance changes against baseline sampling are indicated in green. 
Best results are highlighted in \textbf{bold} and second-best are \underline{underlined}. For all search methods, we use 16 particles. 
\newline { \tiny * The official results from the ICCV 2025 PhysicsIQ Challenge platform are\\ MAGI-1 I2V 37.39 (+7.62) and V2V 62.64 (+7.42).\\ ${}^\dagger$ Sora2 is an API model; guidance ($\nabla$) is not available.}
}
\label{tab:main_physicsiq}
\resizebox{\linewidth}{!}{%
\setlength{\tabcolsep}{1.5pt}
\begin{tabular}{@{}lccccc@{}}
\toprule
& \makecell{Spatial\\ IoU $\uparrow$} 
& \makecell{Spatio-\\temp. IoU $\uparrow$} 
& \makecell{Weighted\\Spatial IoU $\uparrow$} 
& \makecell{MSE $\downarrow$} 
& \makecell{PhysicsIQ\\Score $\uparrow$} \\
\midrule

\multicolumn{6}{c}{\emph{I2V Generation}} \\
\midrule
Sora \citep{sora}     & 0.138 & 0.047 & 0.063 & 0.030 & 10.00 \\
Pika 1.0 \citep{pika_1.0_2023}  & 0.140 & 0.041 & 0.078 & 0.014 & 13.00 \\
SVD  \citep{svd}     & 0.132 & 0.076 & 0.073 & 0.021 & 14.80 \\
Lumiere \citep{lumiere}  & 0.113 & 0.173 & 0.061 & 0.016 & 19.00 \\
VideoPoet \citep{videopoet} & 0.141 & 0.126 & 0.087 & 0.012 & 20.30 \\
Wan2.1 \citep{wan2025}  & 0.153 & 0.100 & 0.112 & 0.023 & 20.89 \\
Gen 3 \citep{runway_gen3_alpha_2024}    & 0.201 & 0.115 & 0.116 & 0.015 & 22.80 \\
Kling1.6 \citep{kling2024}  & 0.197 & 0.086 & 0.144 & 0.025 & 23.64 \\
VLIPP \citep{yang2025vlipp} & N/A & N/A & N/A & N/A & 34.6\phantom{0} \\
\midrule
\textit{vLDM} & 0.221 & 0.120 & 0.144 & \underline{0.008} & 27.76 \\
\quad + \texttt{VideoMAE(BoN)} & 0.229 & 0.132 & 0.152 & 0.009 & 29.42\gain{1.66} \\
\quad + \texttt{Qwen2.5-VL(BoN)} & 0.228 & 0.092 & 0.143 & 0.009 & 26.21\lose{1.55} \\
\quad + \texttt{Qwen3-VL(BoN)} & 0.214 & 0.139 & 0.141 & \underline{0.008} & 28.51\gain{0.75} \\
\rowcolor{rowgray}
\quad + \texttt{WMReward} ($\nabla$)       & 0.221 & 0.122 & 0.144 & 0.009 & 27.88\gain{0.12} \\
\rowcolor{rowgray}
\quad + \bonmethod & 0.249 & 0.155 & \underline{0.170} & \underline{0.008} & 32.90\gain{5.14} \\
\rowcolor{rowgray}
\quad + \guide & 0.234 & 0.175 & 0.165 & \textbf{0.007} & 33.44\gain{5.68} \\
\textit{MAGI-1} \citep{magi1} & 0.252 & 0.146 & 0.151 & 0.011 & 29.77 \\
\quad + \texttt{VideoMAE(BoN)} & 0.251 & 0.151 & 0.153 & 0.012 & 29.95\gain{0.18} \\
\quad + \texttt{Qwen2.5-VL(BoN)} & 0.241 & 0.093 & 0.141 & 0.013 & 24.99\lose{4.78} \\
\quad + \texttt{Qwen3-VL(BoN)} & 0.247 & 0.161 & 0.146 & 0.010 & 30.21\gain{0.44} \\
\rowcolor{rowgray}
\quad + \texttt{WMReward}($\nabla$)     & 0.252 & 0.145 & 0.152 & 0.010 & 29.77\gain{0.00} \\
\rowcolor{rowgray}
\quad + \bonmethod         & 0.253 & \textbf{0.239} & 0.162 & \underline{0.008} & 36.56\gain{6.79} \\
\rowcolor{rowgray}
\quad + \guide*   & 0.267 & 0.218 & 0.168 & \underline{0.008} & 36.28\gain{6.51} \\
\textit{Wan2.2} \citep{wan2025}  & 0.358 & 0.120 & 0.217 & \underline{0.008} & 38.26 \\
\quad + \texttt{VideoMAE(BoN)} & 0.347 & 0.124 & 0.216 & \underline{0.008} & 37.91\lose{0.35} \\
\quad + \texttt{Qwen2.5-VL(BoN)} & 0.356 & 0.116 & 0.213 & 0.009 & 37.58\lose{0.68} \\
\quad + \texttt{Qwen3-VL(BoN)} &  0.353 & 0.128  & 0.216  & \underline{0.008}  & 38.51\gain{0.25} \\
\rowcolor{rowgray}
\quad + \bonmethod & \textbf{0.378} & 0.171 & \textbf{0.247} & \underline{0.008} & \underline{44.39}\gain{6.13} \\
\textit{Sora2} \citep{openai2025sora2} & 0.363 & 0.187 & 0.231 & \textbf{0.007} &  42.30 \\
\quad + \texttt{VideoMAE(BoN)} & 0.365 & 0.193 & 0.234 & \textbf{0.007} & 42.94\gain{0.64} \\
\quad + \texttt{Qwen2.5-VL(BoN)} & 0.340 & 0.166 & 0.210 & \textbf{0.007} & 38.69\lose{3.61} \\
\quad + \texttt{Qwen3-VL(BoN)} & 0.349& 0.206 & 0.226 & \textbf{0.007} &  42.54\gain{0.24}\\ 
\rowcolor{rowgray}
\quad + \bonmethod$^\dagger$         & \underline{0.377}	& \underline{0.229} & \underline{0.246} & \textbf{0.007} & \textbf{46.43}\gain{4.13} \\
\midrule
\multicolumn{6}{c}{\emph{V2V Generation}} \\
\midrule
Lumiere \citep{lumiere}  & 0.170 & 0.155 & 0.093 & 0.013 & 23.00 \\
VideoPoet \citep{videopoet} & 0.204 & 0.164 & 0.137 & 0.010 & 29.50 \\
\midrule
\textit{MAGI-1} \citep{magi1}  & 0.416 & 0.279 & 0.294 & \textbf{0.005} & 55.22 \\
\quad + \texttt{VideoMAE(BoN)} & 0.407 & 0.286 & 0.285 & \textbf{0.005} & 54.71\lose{0.51} \\
\quad + \texttt{Qwen2.5-VL(BoN)} & 0.415 & 0.272 & 0.293 & \textbf{0.005} & 54.71\lose{0.51} \\
\quad + \texttt{Qwen3-VL(BoN)} & 0.407 & 0.279 & 0.289 & \textbf{0.005} & 54.42\lose{0.80} \\
\rowcolor{rowgray}
\quad + \texttt{WMReward} ($\nabla$) & 0.414 & 0.279 & 0.292 & \textbf{0.005} & 55.02\lose{0.20} \\
\rowcolor{rowgray}
\quad + \bonmethod & \underline{0.435} & \underline{0.324} & \underline{0.316} & \textbf{0.005} & \underline{60.34}\gain{5.12} \\
\rowcolor{rowgray}
\quad + \guide* & \textbf{0.439} & \textbf{0.339} & \textbf{0.325} & \textbf{0.005} & \textbf{62.00}\gain{6.78} \\

\bottomrule

\end{tabular}%
}
\end{table}

\Cref{tab:main_physicsiq} demonstrates that sampling with \method consistently outperforms vanilla sampling for vLDM, MAGI-1, and Sora2, achieving state-of-the-art performance (best and second-best) across baselines and evaluation dimensions.
In particular, by using the appropriate sampling strategy, \texttt{$\nabla$+BoN}, we achieve $62.0\%$ final PhysicsIQ score on V2V generation, which significantly outperforms the previous state-of-the-art MAGI-1 model by $6.78\%$.
For I2V, sampling with \method surpasses the previous state-of-the-art Sora2~\citep{openai2025sora2} by $4.13\%$ on PhysicsIQ.\looseness-1

\vspace{-2mm}

\paragraph{Comparing Reward Signals.}
We compare \method against alternative physics plausibility signals based on foundation models such as VideoMAE and Qwen VL (2.5-7B-Instruct and 3-8B-Instruct). For VideoMAE~\citep{tong2022videomae}, a self-supervised video masked autoencoder that learns spatiotemporal representations by reconstructing masked patches in pixel space, we use a surprise score given by the reconstruction error between the predicted pixels and the generated pixels as reward signal~\citep{garrido2025intuitive}. For Qwen2.5-VL~\citep{bai2025qwen2} and Qwen3-VL~\citep{yang2025qwen3}, following~\citet{jang2025dreamgen}, we pose a physics-plausibility question and request a binary 0/1 answer, then follow practice in \citep{lin2024vqascore} and use the corresponding positive token logit as the reward. 

As shown in \Cref{tab:main_physicsiq} and \Cref{fig:teaser2}, \method yields stronger performance than alternative signals, indicating that predictive latent-space surprise is a more effective proxy for physics plausibility than pixel reconstruction or VLM judgment, which aligns with recent work suggesting that latent world models exhibit better physics understanding~\citep{garrido2025intuitive}. With these results, we show that the physics knowledge in latent world models may be transferred as reward signal to improve video generation.\looseness-1

\begin{table}[t]
\centering
\caption{\textbf{Human evaluation results on PhysicsIQ and VideoPhy.} Pairwise human preference judgments across three criteria: Physics Plausibility, Visual Quality, and Prompt Alignment. Winning rate (Win) is the fraction of non-neutral comparisons won, $\frac{\text{wins}}{\text{wins} + \text{losses}} \times 100$. Accuracy is computed as $\frac{\text{wins} + 0.5\times\text{neutrals}}{\text{total}} \times 100$ to account for tie. Higher is better, best in bold.
}
\label{tab:human_eval_physicsiq}
\setlength{\tabcolsep}{3pt}
\resizebox{\linewidth}{!}{
\begin{tabular}{@{}lcccccccc@{}}
\toprule
& \multicolumn{2}{c}{\makecell{\textbf{Physics}\\\textbf{Plausibility}}}
& \multicolumn{2}{c}{\makecell{\textbf{Visual}\\\textbf{Quality}}}
& \multicolumn{2}{c}{\makecell{\textbf{Prompt}\\\textbf{Alignment}}}
& \multicolumn{2}{c}{\textbf{Overall}} \\
\cmidrule(lr){2-3} \cmidrule(lr){4-5} \cmidrule(lr){6-7} \cmidrule(lr){8-9}
& Win & Acc.
& Win & Acc.
& Win & Acc.
& Win & Acc. \\

\midrule
\multicolumn{9}{c}{\textit{PhysicsIQ}} \\
\midrule
\textit{vLDM} & 46.9 & 45.2 & 45.3 & 41.2 & 48.7 & 46.8 & 47.0 & 44.3 \\
\rowcolor{rowgray}\quad + \guide & \textbf{53.1} & \textbf{54.8} & \textbf{54.7} & \textbf{58.8} & \textbf{51.3} & \textbf{53.2} & \textbf{53.0} & \textbf{55.7} \\
\textit{MAGI-1} & 45.1 & 41.3 & 47.1 & 42.9 & 44.3 & 35.1 & 45.5 & 40.0 \\
\rowcolor{rowgray}\quad + \guide & \textbf{54.9} & \textbf{58.7} & \textbf{52.9} & \textbf{57.1} & \textbf{55.7} & \textbf{64.9} & \textbf{54.5} & \textbf{60.0} \\

\midrule
\multicolumn{9}{c}{\textit{VideoPhy}} \\
\midrule
\textit{vLDM} & 43.8 & 38.0 & 47.4 & 41.9 & \textbf{51.0} & \textbf{53.3} & 47.4 & 43.2 \\
\rowcolor{rowgray}\quad + \guide & \textbf{56.2} & \textbf{62.0} & \textbf{52.6} & \textbf{58.1} & 49.0 & 46.7 & \textbf{52.6} & \textbf{56.8} \\
\textit{MAGI-1} & 40.7 & 30.4 & 42.8 & 33.3 & 49.5 & 48.6 & 44.3 & 36.8 \\
\rowcolor{rowgray}\quad + \guide & \textbf{59.3} & \textbf{69.6} & \textbf{57.2} & \textbf{66.7} & \textbf{50.5} & \textbf{51.4} & \textbf{55.7} & \textbf{63.2} \\ 

\bottomrule
\end{tabular}
}
\end{table}

\vspace{-2mm}

\paragraph{Human Study.} We also supplement our evaluations with a human study to verify the effectiveness of \method. We conduct a human study on the full PhysicsIQ benchmark with a side-by-side comparison interface where five annotators view pairs of generated videos along with the original text prompt and additional conditioning information. 
For each video pair, annotators provide judgments across three criteria: Physics Plausibility, Visual Quality, and Prompt Alignment (further details in Appendix
\Cref{apx:human_details}). We obtain two groups of 198 annotations for both vLDM (I2V) and MAGI-1 (V2V) generations, comparing the baseline vanilla sampling strategy against \guide. For each criterion, the annotators select their preference among the shown videos or mark them as neutral. Results are aggregated using win rates (excluding neutrals) and accuracy scores to account for ties ((wins + 0.5 × neutrals) / total).
\Cref{tab:human_eval_physicsiq} demonstrates that \method delivers significant improvement in all three evaluation criteria, with the winning rate in physics plausibility being most remarkable.\looseness-1

\subsection{Text-conditioned Generation}
For evaluating the T2V setup, we rely on VideoPhy \citep{bansal2024videophy}. Following practice in \citet{zhang2025videorepa} and \citet{ wang2025wisa}, we adopt the automatic VLM-based evaluator, which queries a VLM with templated questions to score semantic adherence (SA) and physics consistency (PC) on a scale of 1-5 for each generated video. We report the per-axis pass rates with each metric larger than 4 over the 344 benchmark prompts. For all methods, we use 8 particles for search.\looseness-1 

As shown in \Cref{tab:videophy_main}, incorporating \method into sampling substantially improves the physics consistency of both MAGI-1 and vLDM by $8.1\%$ and $6.9\%$, respectively, surpassing all baseline models. 
We also note that the semantic adherence decreases potentially due to the VJEPA surprise not including semantic information from the text-condition, and to further verify the trade-off between semantic adherence and physics consistency we perform a human study.\looseness-1

\begin{table}[t]
\centering
\caption{\textbf{Text-conditioned Generation Performance on VideoPhy.} Physics Consistency (PC) and Semantic Adherence (SA) are computed with a VLM-based evaluator that measures how well generated videos follow physics laws and text prompts. Higher is better, best is bold. For all search methods, we use 8 particles.
}
\setlength{\tabcolsep}{3pt}
\resizebox{\linewidth}{!}{%
\begin{tabular}{@{}lcccccccc@{}}
\toprule
& \multicolumn{2}{c}{Solid-Solid} 
& \multicolumn{2}{c}{Solid-Fluid} 
& \multicolumn{2}{c}{Fluid-Fluid} 
& \multicolumn{2}{c}{Overall} \\
\cmidrule(lr){2-3}\cmidrule(lr){4-5}\cmidrule(lr){6-7}\cmidrule(lr){8-9}
 & SA & PC & SA & PC & SA & PC & SA & PC \\
\midrule
VideoCrafter2 \citep{chen2024videocrafter2}       & 50.4 & 32.2 & 50.7 & 27.4 & 48.1 & 29.1 & 50.3 & 29.7 \\
DreamMachine  \citep{luma2024dreammachine}        & 55.1 & 21.7 & 59.6 & 23.3 & 58.2 & 18.2 & 57.5 & 21.8 \\
LaVIE \citep{wang2025lavie}                & 40.8 & 18.3 & 48.6 & 37.0 & 69.1 & 50.9 & 48.7 & 31.5 \\
HunyuanVideo \citep{kong2024hunyuanvideo}    & 55.2 & 16.1 & 67.1 & 30.1 & 54.5 & 54.5 & 60.2 & 28.2 \\
\midrule
\textit{vLDM} & 46.6 & 20.7 & 67.5 & 28.6 & 52.4 & 48.6 & 56.9 & 28.0 \\
\rowcolor{rowgray} \quad + \bonmethod            & 42.7 & \textbf{28.7} & 59.6 & 36.3 & 38.5 & \textbf{50.0} & 50.3 & \textbf{34.9} \\
\rowcolor{rowgray} \quad + \guide   & 39.2 & 25.2 & 65.1 & \textbf{37.0} & 50.0 & \textbf{50.0} & 53.5 & 34.3 \\
\midrule
\textit{Wan2.2} \citep{wan2025} &52.7	&13.9&	71.9&	\textbf{17.8}&	67.5	&\textbf{30.2}	&63.5	&18.1 \\
\rowcolor{rowgray} \quad + \bonmethod & 43.7 & \textbf{21.8}& 63.8 & 15.9 & 53.8 & 26.9 & 55.2 & \textbf{20.9} \\
\midrule
\textit{MAGI-1} \citep{magi1} & 42.2 & 19.0 & 67.2 & 27.7 & 51.9 & 33.5 & 54.4 & 25.0 \\
\rowcolor{rowgray} \quad + \bonmethod            & 30.8 & 28.0 & 59.6 & \textbf{36.3} & 45.5 & 36.4 & 45.3 & 32.8 \\
\rowcolor{rowgray} \quad + \guide   & 29.4 & \textbf{28.7} & 61.6 & 30.1 & 40.0 & \textbf{52.7} & 44.8 & \textbf{33.1} \\
\bottomrule
\end{tabular}}
\label{tab:videophy_main}
\end{table}

\vspace{-2mm}

\paragraph{Human Study.}
To complement the VLM-based evaluation, we conduct a human study following the same protocol in \Cref{sec:physicsiq} on a subset of 100 VideoPhy generations. As shown in \Cref{tab:human_eval_physicsiq}, human judgement highlights improvements in both physics plausibility and visual quality when using \method for T2V generation. The slight decrease in prompt alignment for vLDM is consistent with the text-agnostic nature of the VJEPA's surprise reward, but the drop is small relative to the gains, yielding a net improvement in the overall metric. We expect this limitation can be mitigated by developing compositional or text-conditioned physics plausibility rewards, an interesting direction for future work.\looseness-1

\subsection{Scaling the Search Space of \textbf{\method}}
\label{sec:scaling}

\begin{figure*}[htbp]
    \centering
    \begin{minipage}[t]{0.46\textwidth}
        \centering
        \includegraphics[width=\linewidth]{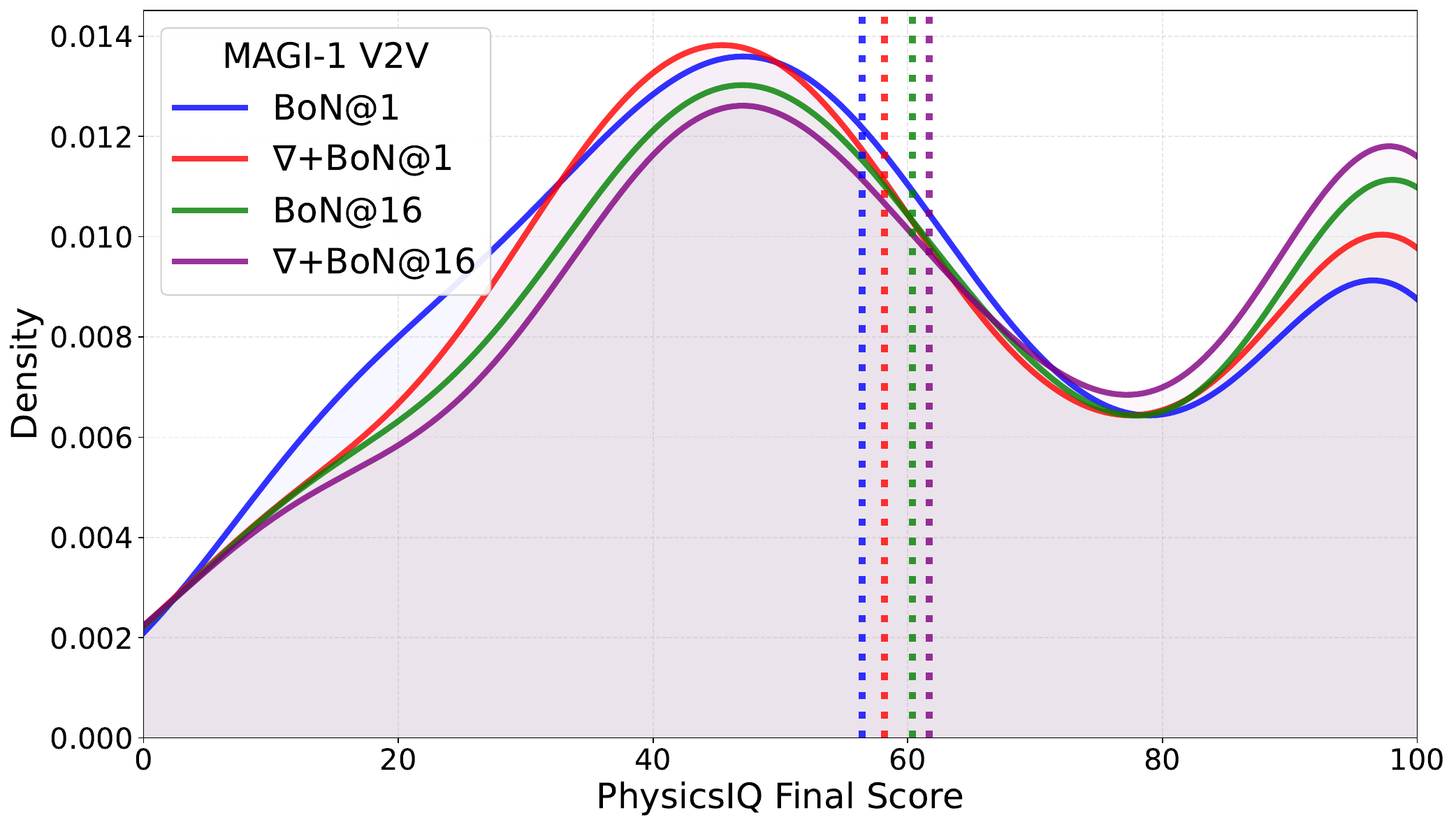}
    \end{minipage}
    \begin{minipage}[t]{0.46\textwidth}
        \centering
        \includegraphics[width=\linewidth]{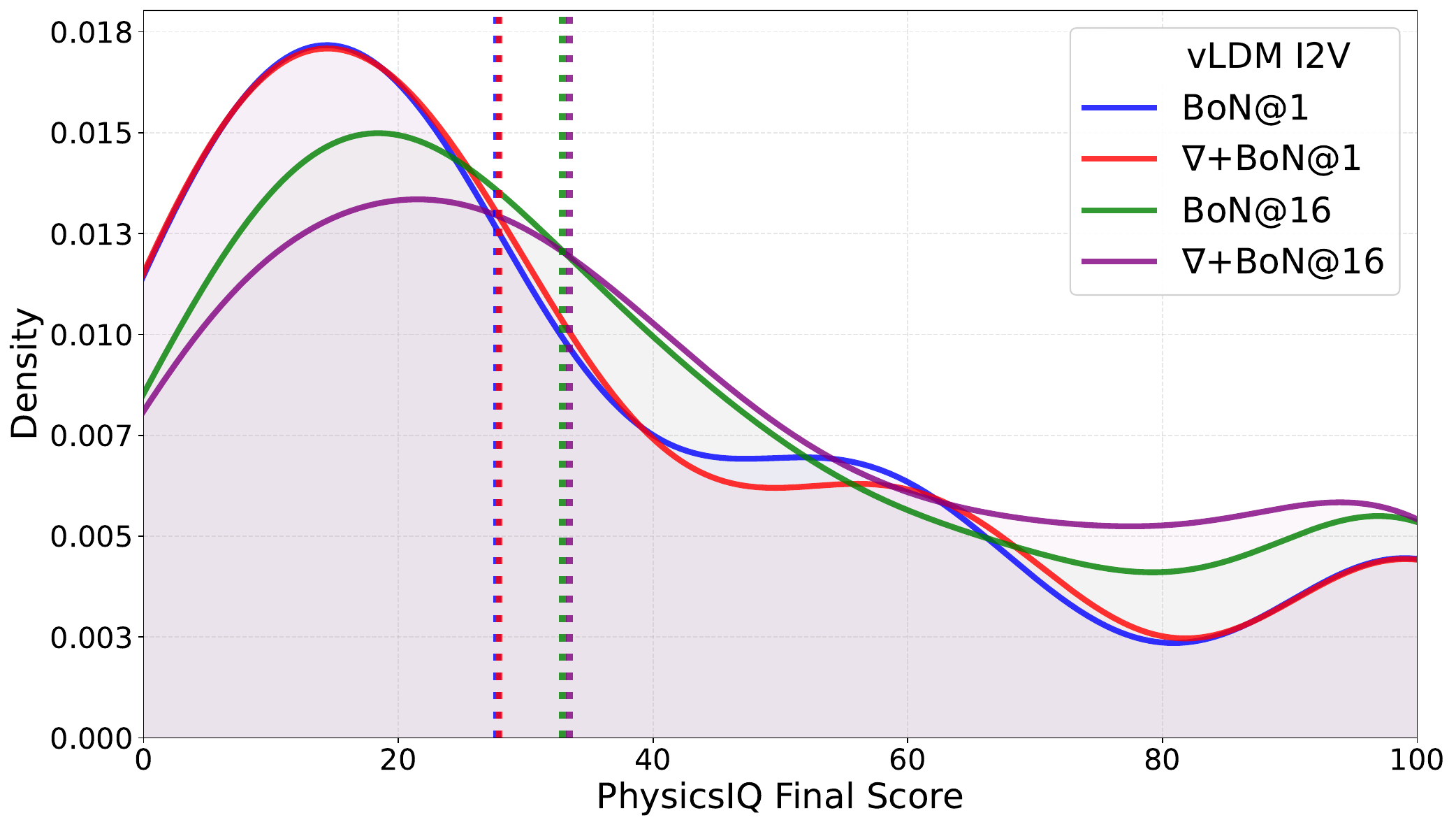}
    \end{minipage}
    \captionsetup{aboveskip=2pt,belowskip=0pt} 
    \caption{\textbf{Improving Performance via Particle Scaling and Guidance.} Visualization of PhysicsIQ score distributions with a Gaussian KDE for MAGI-1 V2V (left) and vLDM I2V (right) generations. Scaling the number of particles (here, 1 vs.\ 16) yields substantial gains for Best-of-$N$, and adding guidance further sharpens the distribution toward higher physics plausibility. This leads to overall higher average score shown as dashed vertical line. 
    }
    \label{fig:distribution}
\end{figure*}

\vspace{0.2em}
We now study the scaling behavior of \method with the size of the search space. In particular, we scale the number of particles and display results in \Cref{fig:teaser2} (right). We observe that, by scaling up the number of particles, there is a steady improvement of PhysicsIQ score. The performance boost is most significant for $N\!\leq\!4$, and steadily increase for larger $N$.
Additionally, as the number of particles increases, the performance variance is smaller and gradually stabilizes. Moreover, as shown in \Cref{fig:distribution}, allocating more compute by increasing the number of particles and applying guidance steers the PhysicsIQ score distribution to concentrate in the high-score region. 
The shift of distributions from $N\!=\!1$ to $N\!=\!16$ shows the effectiveness of \bonmethod by using more particles, while \guide with guidance further sharpening the upper tail of the distribution, confirming its stronger scaling behavior. The consistent scaling behavior of \method on PhysicsIQ suggests that by increasing inference compute allocation, we improve the physical plausibility of video generation.\looseness-1

\begin{table}[t]
\centering
\caption{\textbf{Computation Cost.} For each model, the first row shows absolute values (runtime in seconds, memory in GB per GPU). Subsequent rows show overhead multipliers relative to baseline. $N$ is the search budget (number of particles). All experiments on H200 GPUs; MAGI-1 uses 8 GPUs, vLDM uses 1 GPU.}
\label{tab:compute_min}
\setlength{\tabcolsep}{3pt}
\resizebox{\linewidth}{!}{
\begin{tabular}{@{}lccr}
\toprule
Method & Time & Memory & $N$ \\
\midrule
\textit{vLDM (baseline)}         & 106.77 ± 0.43 s & 27.64 GB & 1 \\
\rowcolor{rowgray}\quad + \bonmethod                    & $\times N\phantom{0.00}$ & $\times 1.00$ & 8 \\
\rowcolor{rowgray}\quad + \texttt{WMReward}($\nabla$)   & $\times 5.02$\phantom{N} & $\times 4.27$ & 1 \\
\rowcolor{rowgray}\quad + \guide                        & $\times 5.02N$ & $\times 4.27$ & 8 \\
\midrule
\textit{MAGI-1 (baseline)}            & 265.66 ± 4.49 s & 50.03 GB & \phantom{0}1 \\
\rowcolor{rowgray}\quad + \bonmethod                    & $\times N\phantom{0.00}$ & $\times 1.00$ & 16 \\
\rowcolor{rowgray}\quad + \texttt{WMReward}($\nabla$)   & $\times 4.96\phantom{N}$ & $\times 2.07$ & \phantom{0}1 \\
\rowcolor{rowgray}\quad + \guide                        & $\times 4.96N$ & $\times 2.07$ & 16 \\
\bottomrule
\end{tabular}
}
\vspace{-3mm}
\end{table}

\vspace{-2mm}
\paragraph{Runtime and Memory.}
Inference-time alignment methods trade additional computation for improved performance---a recent paradigm shift, from long reasoning traces to test-time search in general. In \Cref{tab:compute_min}, we show the computational overhead induced by \method. \bonmethod is parallelized as $N$ vanilla sampling trajectories with the same memory bound, and its memory footprint scales linearly with the number of particles. Guidance introduces extra computation mainly from the gradient backpropagation, and is proportional to the number of steps where guidance is performed through the denoising process. One can choose whether to incorporate first-order information of reward signal as budget allows. In general, \method can be adapted to the available computation budget with appropriate sampling methods, offering a spectrum of compute-performance tradeoffs.\looseness-1

\subsection{Analysis and Ablation}
Having established that \method achieves strong improvements in physics plausibility and scales effectively with computational budget, we now investigate several key questions about the method's properties.
\vspace{-3mm}

\paragraph{Does \textbf{\method} impair perceptual qualities?}
To assess whether physics improvements come at the cost of general perceptual quality, we use the VBench evaluators \citep{huang2024vbench} to rate generated PhysicsIQ and VideoPhy videos across six key visual quality metrics: subject consistency, background consistency, motion smoothness, temporal flickering, imaging quality, and aesthetic quality. \looseness-1
\begin{table}[t]
    \centering
    \caption{\textbf{General Visual Quality with VBench Evaluators.} Visual quality metrics computed on videos generated for the PhysicsIQ and VideoPhy datasets. Higher is better for all metrics.}
    \setlength{\tabcolsep}{2pt}
    \resizebox{\linewidth}{!}{
    \begin{tabular}{@{}lcccccc@{}}
        \toprule
        & \makecell[c]{\textbf{Subject} \\ \textbf{Consistency}} 
        & \makecell[c]{\textbf{Background} \\ \textbf{Consistency}} 
        & \makecell[c]{\textbf{Motion} \\ \textbf{Smoothness}} 
        & \makecell[c]{\textbf{Temporal} \\ \textbf{Flickering}} 
        & \makecell[c]{\textbf{Imaging} \\ \textbf{Quality}} 
        & \makecell[c]{\textbf{Aesthetic} \\ \textbf{Quality}} \\
        \midrule
        \multicolumn{7}{c}{\textit{PhysicsIQ}} \\
        \midrule
        
        \textit{vLDM} & 94.68 & 95.60 & 99.57 & 99.48 & 67.47 & 48.94 \\
        \rowcolor{rowgray} \quad + \texttt{WMReward}($\nabla$)   & 94.64 & 95.58 & 99.56 & 99.47 & 67.50 & 48.95 \\
        \rowcolor{rowgray} \quad + \bonmethod & 95.29 & 95.89 & 99.62 & 99.61 & 67.71 & 49.19 \\
        
        \rowcolor{rowgray} \quad + \guide & 95.25 & 95.89 & 99.62 & 99.61 & 67.83 & 49.03 \\
     
        \textit{MAGI-1} & 95.82 & 97.25 & 99.70 & 99.82  & 69.07 & 47.05 \\
        \rowcolor{rowgray} \quad + \texttt{WMReward}($\nabla$)       & 95.90 & 97.25 & 99.70 & 99.82  & 69.15 & 47.05 \\
        \rowcolor{rowgray} \quad + \bonmethod & 95.87 & 97.30 & 99.71 & 99.84  & 69.24 & 46.88 \\ 
        \rowcolor{rowgray} \quad + \guide & 95.88 & 97.30 & 99.71 & 99.85  & 69.25 & 46.94 \\ 
        \addlinespace
        
        \midrule
        \multicolumn{7}{c}{\textit{VideoPhy}} \\
        \midrule
        \textit{vLDM}         & 93.73 & 95.79 & 98.57 & 97.80 & 60.34 & 49.71 \\
        \rowcolor{rowgray} \quad + \texttt{WMReward}($\nabla$)       & 93.72 & 95.78 & 98.57 & 97.80 & 60.05 & 49.69 \\
        \rowcolor{rowgray} \quad + \bonmethod            & 94.82 & 96.31 & 98.69 & 98.02 & 59.44 & 49.54 \\
        \rowcolor{rowgray} \quad + \guide   & 94.67 & 96.36 & 98.72 & 98.13 & 60.03 & 49.87 \\
        \textit{MAGI-1} & 94.43 & 95.87 & 98.80 & 98.08 & 56.15 & 46.39 \\
        \rowcolor{rowgray} \quad + \texttt{WMReward}($\nabla$)  & 94.58 & 95.92 & 98.81 & 98.12 & 56.17 & 46.46 \\
        \rowcolor{rowgray} \quad + \bonmethod & 96.27 & 96.68 & 98.92 & 98.44 & 58.15 & 47.54 \\
        \rowcolor{rowgray} \quad + \guide & 96.54 & 96.87 & 98.97 & 98.54 & 57.84 & 47.73 \\

        \bottomrule
    \end{tabular}}
    \label{tab:vbench}
\end{table}

As shown in \Cref{tab:vbench}, image quality and aesthetic quality tend to show small improvements, possibly due to the stronger physics plausibility achieved by \method. This trend aligns with the human-preference results in \Cref{tab:human_eval_physicsiq}. We also observe improved temporal consistency, motion smoothness, and temporal flickering, suggesting that by following physics principles, such as artifact suppression, and adherence to mass conservation and continuity, the overall perceptual quality of generated videos are also improved.\looseness-1
\vspace{-3mm}

\paragraph{How does \textbf{\method} perform with other sampling methods?}

\begin{table}[t]
\centering
\footnotesize
\caption{\textbf{Comparison of Sampling Methods.} vLDM I2V PhysicsIQ generation under different sampling schemes. Columns indicate number of particles ($N$), showing scaling effects.\looseness-1}
\label{tab:alignment-comparison}
\begin{tabular}{@{}lccccc@{}}
\toprule
  & 2 & 4 & 8 & 16 \\
\midrule
\texttt{WMReward(SMC)}           & $27.14$ & $26.96$ & $28.12$ & $29.24 $ \\
\texttt{WMReward(SVDD)}          & $27.31 $ & $28.57$ & $28.54$ & $28.87$ \\
\bonmethod             & $27.76$ & $\mathbf{31.87}$ & $32.32$ & $32.90$ \\
\guide    & $\mathbf{27.88}$ & $31.05$ & $\mathbf{32.62}$ & $\mathbf{33.44}$ \\
\bottomrule
\end{tabular}
\vspace{-5mm}
\end{table}

In principle, one can use any valid sampling schemes to sample from the \method tilted distribution. We further demonstrate the compatibility of the proposed reward by alternative sampling scheme with SMC \citep{singhal2025smc} and SVDD \citep{li2024svdd} (detailed in \Cref{sec:background}) using vLDM on the PhysicsIQ benchmark. Empirically, both SMC and SVDD deliver gains over vanilla sampling, yet, they still underperform relative to \texttt{BoN} and $\nabla$ + \texttt{BoN}, as demonstrated in \Cref{tab:alignment-comparison}. Intuitively, because VJEPA-2 surprise is differentiable, its gradients provide step-wise guidance that plays a role similar to, and often more accurate than, the multi-rollout importance estimates used by SVDD. Also, SMC shows smaller gains, likely because early-stage reward estimates are noisy, causing resampling to collapse onto suboptimal seeds. Practically, SMC also requires multiple parallel denoising trajectories per sample for resampling, which increases the memory footprint and is less trivial to parallelize than BoN search. For these reasons, we adopt \guide as a more practically effective sampling scheme.\looseness-1
\vspace{-3mm}

\paragraph{How robust is the \method to VJEPA size and hyper-parameters?}
One key design factor in our experiments is the choice of hyperparameters for the VJEPA surprise reward \(r(\cdot)\) in \Cref{eq:surprise}. We study how robust this reward is when transferring physics understanding to video generation. As shown in \Cref{tab:vjepa_design_robustness}, such transfer remains relatively stable across the context length \(C\), prediction horizon \(M\), and stride \(s\). We also observe that physics plausibility gains scale with the size of the reward model (from \texttt{ViT-huge} to \texttt{ViT-giant}), suggesting that stronger VJEPA backbones yield better performance without any fine-tuning of the underlying video generator.

\begin{table}[t]
        \centering
        \caption{
            \textbf{Robustness of V-JEPA Surprise Design.} Comparison showing VJEPA architecture and reward hyperparameter against \bonmethod results. We report results using 16 particles. 
        }
        \resizebox{\linewidth}{!}{
        \begin{tabular}{@{}cccccccc@{}}
            \toprule
            \textbf{Arch Size}  & \textbf{Window} & \textbf{Context} & \textbf{Stride} & \textbf{FPS} & \textbf{Final Score}$\uparrow$ \\
            \midrule
        ViT-giant & 32 & 16 & 16 & 24 & \textbf{60.78} \\
        ViT-giant & 16 & 8 & 8 & 16 & 60.34 \\
        ViT-giant   & 16 & 8 & 8  & 24 & 60.09 \\
        ViT-giant   & 32 & 16 & 8  & 24 & 60.05 \\
        ViT-huge   & 48 & 24 & 8 & 24 & 60.04 \\
        ViT-giant   & 16 & 8 & 4  & 24 & 59.91 \\
        ViT-huge   & 16 & 8 & 8 & 24 & 59.77 \\
        ViT-huge   & 16 & 8 & 4 & 24 & 59.62 \\
        ViT-huge   & 32 & 16 & 16 & 24 & 57.84 \\
        ViT-huge   & 32 & 16 & 8 & 24 & 57.09 \\
        \bottomrule
    \end{tabular}
    }
    \label{tab:vjepa_design_robustness}
\end{table}

\section{Related Work}
\label{sec:background}
\paragraph{Video Diffusion Models.}
Diffusion and flow-based video generative models~\citep{magi1, sora, openai2025sora2,videopoet, lumiere, wan2025} learn to generate video sequences by reversing a noising process on temporal visual data. While they achieve promising visual quality in generation, they still often produce physically implausible videos~\citep{bansal2024videophy, bansal2025videophy2,physicsIQ, yuan2025likephys}.
In order to mitigate this shortcoming, there has been substantial work devising pre-training and post-training approaches focused on explicitly injecting physics constraints~\citep{yuan2025newtongen}, enabling motion prediction~\citep{chefer2025videojam}, enhancing dynamics~\citep{li2024generativedynamics} or leveraging learned physics priors from foundation models through model finetuning or distillation~\citep{cao2024teaching, zhang2025videorepa,zhang2025think}. Yet, another relatively underexplored line of work has focused on devising inference-time methods to improve physics. In particular,~\citet{xue2025physt2v} explore the manifold learned by the generative model through prompt rewriting and~\citet{yang2025vlipp} leverage VLMs to perform motion planning on top of a motion-controllable video generative model~\citep{burgert2025go}.

\textbf{Physics Understanding in Vision Models.}
Physics understanding is essential to a model’s ability to reason and predict scenes under physical laws~\citep{bear2021physion,riochet2018intphys,bordes2025intphys}, and in particular, it is important to ensure physics plausibility in video generation~\citep{physicsIQ,yuan2025likephys}. Recent work~\citep{bordes2025intphys} has investigated the intuitive physics understanding of vision models, including VLMs~\citep{bai2025qwen2, comanici2025gemini}, generative models~\citep{agarwal2025cosmos}, and self-supervised learning (SSL) models~\citep{tong2022videomae, wang2023videomae2,vjepa2}, showing that video SSL approaches based on the Joint-Embedding-Predictive-Architecture (JEPA)~\citep{bardes2024vjepa,vjepa2} exhibit the highest performance among predictive methods. Moreover, intuitive physics properties such as object permanence and shape consistency have been shown to emerge in the video JEPA (VJEPA) models~\citep{garrido2025intuitive}. VJEPA consists of a video encoder and a predictor trained on Internet-scale data with a masked predictive objective in the representation space. Predicting the future in the representation space reduces the sensitivity to appearance details and holds the potential to emphasize other aspects such as dynamics and interaction.\looseness-1 

\textbf{Inference-Time Alignment for Image Diffusion Models.}
Inference-time alignment steers a pre-trained diffusion model toward desired properties during generation without retraining. On the one hand, there are derivative-free approaches~\citep{singhal2025smc,wu2023practical,lee2025adaptive,li2024svdd,li2025dsearch,jain2025diffusion,park2025steerx}. Among those, BoN selects the highest-scoring candidates at the end of the denoising process~\citep{ma2025inferencescaling}. SMC-based approaches~\citep{singhal2025smc,wu2023practical} assign importance weights to multiple denoising trajectories and perform resampling during generation. Value-based importance sampling methods, such as SVDD~\citep{li2024svdd}, steer a single trajectory by resampling according to the local step reward.  
On the other hand, we have gradient-based approaches that adjust the score function~\citep{song2020score,bansal2023universal, ye2024tfg,hemmat2023feedback,askari2024cvs,dall2025chamfer,askari2025dp}, steering the denoising process toward high reward regions-- resulting in \eg, high utility or high diversity samples. These assume access to differentiable reward functions. Despite their different implementations, these methods share a similar principle of using reward signals to refine local steps or enable better global exploration.\looseness-1

\section{Conclusion and Discussion}
\label{sec:concl}
In this work, we presented \method, an inference-time alignment method to improve physics plausibility of video generative models, and highlighted the viability of using latent world models (\ie, VJEPA-2) as reward signal. Throughout our experimentation, we demonstrated the potential of leveraging latent world model's to improve physics in video generation, yielding substantial improvements on the challenging PhysicsIQ and VideoPhy benchmarks without requiring any further training.\looseness-1

There are three potential avenues for improving this line of work. (1) \emph{Improving video generative models}: The quality of video generative model determines the set of potential solutions that can be explored with our method; developing stronger and more searchable models is a promising avenue. (2) \emph{Improving reward models:} The strength of the reward model heavily influences the overall performance of the system. Although VJEPA-2 is already informative, its training data is currently limited and as a result the model is not able to cover all physics phenomena, \eg, we observe that material understanding such as weight and friction is limited. Future work on latent world models could provide broader understanding of physical phenomena. (3) \emph{Improving search algorithms:} Better and more efficient search algorithms offer a promising avenue to improve performance. Early in the diffusion process, the intermediate estimated video samples $x_{0|t}$ are blurry, potentially leading to unreliable predictions in the reward.\looseness-1

\clearpage
\newpage
\bibliographystyle{assets/plainnat}
\bibliography{main}

\clearpage
\newpage
\beginappendix

\section{Implementation Details}
\label{apx:implementation_details}
In the following, we touch on the implementation details of \method, including generation settings and adaptation to different generation paradigms.
\subsection{Generation Settings}
\label{apx:generation_settings}
For all experiments, we use a vLDM transformer with a spatiotemporal VAE for compression and text–video alignment, and MAGI-1-24B~\citep{magi1}, an autoregressive diffusion video model that generates videos chunk-by-chunk using block-causal attention for long-horizon consistency. Their corresponding generation hyperparameters are as follows.

\begin{table}[h]
\vspace{-2mm}
\centering
\caption{\textbf{Generation hyperparameters.}}
\label{tab:gen_hparams_transposed}
\resizebox{\linewidth}{!}{
\begin{tabular}{@{}lccccc@{}}
\toprule
& \multicolumn{2}{c}{\textbf{\textit{VideoPhy}}} & \multicolumn{3}{c}{\textbf{\textit{PhysicsIQ}}} \\
\cmidrule(lr){2-3} \cmidrule(lr){4-6}
\textbf{Hyperparameter} & MAGI-1 & vLDM & MAGI-1 (I2V) & MAGI-1 (V2V) & vLDM \\
\midrule
Height  & 480 & 480 & 720 & 720 & 480 \\
Width  & 832 & 720 & 1280 & 1280 & 720 \\
Number of frames & 48 & 49 & 120 & 120 & 49 \\
FPS & 24 & 8 & 24 & 24 & 8 \\
Number of steps & 16 & 50 & 32 & 32 & 50 \\
CFG scale & 7.5 & 6.0 & 7.5 & 7.5 & 6.0 \\
Context guidance scale & 1.5 & -- & 1.5 & 1.5 & -- \\

guidance frequency & 3 & 3 & 5 & 3 & 1 \\
VJEPA guidance scale & 0.005 & 0.003 & 0.005 & 0.005 & 0.001 \\
\bottomrule
\end{tabular}}
\end{table}

For generation resolution, FPS, we use the recommended settings from the official video generative model repository. Also, for CFG and other guidance implemented in MAGI-1, we follow the default settings in official codebase. The number of generated frames varies according to the specification of the evaluation dataset. For PhysicsIQ, the generated video is required to be exactly 5 seconds. Thus, we generate 49 frames with vLDM and trim them to 40 frames under 8 FPS. For MAGI-1, we generate 120 frames under 24 FPS. For VideoPhy, while there is no explicit requirement on duration and number of frames, we follow the paradigm in the official code to generate relative short video clips with the specification shown.
The guidance frequency indicates the time-step interval in which we apply guidance during the denoising process.
For vLDM, we use a DDIM~\citep{song2020denoising} scheduler. For MAGI-1, we use standard linear rectified flow sampler. For MAGI-1, we use the distilled 24B checkpoint, and run the inference on 8 H200 GPUs in parallel. For vLDM, we run the inference on a single H200 GPU. Also, for sampling with \method, we use a VJEPA2 ViT-giant model. The input frame size is $256 \times 256$. We choose window size, context length, and stride to be 16, 8, 8 for all experiments.\looseness-1

\subsection{Adaptation to Different Generation Paradigms}
Current video diffusion models follow two predominant paradigms: holistic generation~\citep{wan2025,agarwal2025cosmos}, which denoises all frames simultaneously at the same noise level, and autoregressive generation~\citep{magi1}, which generates videos sequentially in temporal chunks, each with its own noise schedule. We implement \method on models based on both paradigms (vLDM and MAGI-1). Practically, in both cases, the BoN search implementation is the same, while the implementation of guidance ($\nabla$) varies.
For holistic generation, we use the the combined CFG and \method guidance in \Cref{eq:score_holistic}.
\begin{equation}
\label{eq:score_holistic}
\hspace*{-0.3em}
\begin{aligned}
\nabla_{x_t} \log p_{\lambda}\bigl(x_t \mid \textrm{txt}\bigr)
&= (1 - \omega_{\textrm{txt}})\, \nabla_{x_t} \log p(x_t) \\
&\quad + \omega_{\textrm{txt}} \, \nabla_{x_t} \log p\bigl(x_t \mid \textrm{txt}\bigr) \\
&\quad - \omega_s \nabla_{x_t} r\bigl(x_t \mid \textrm{txt}\bigr).
\end{aligned}
\end{equation}
Specifically, we adopt a sliding window approach to split the video into context and prediction target chunks, compute the VJEPA surprise on each window, and average over the whole sequence.
For autoregressive generation, we perform guidance as follows:\looseness-1
\begin{equation}
\label{eq:score_autogressive}
\hspace*{-0.3em}
\begin{aligned}
        \nabla_{x_t} \log p_{\lambda} \bigl(x_t \mid x^{<k}_{t}, \textrm{txt}\bigr) = (1- \omega_{<k}) \nabla_{x_t} \log p(x_t) \\ + \  (\omega_{<k} - \omega_{\textrm{txt}}) \nabla_{x_t} \log p\bigl(x_t \mid x^{<k}_{t}\bigr) \\ + \ \omega_{\textrm{txt}} \nabla_{x_t} \log p\bigl(x_t \mid x^{<k}_{t}, \textrm{txt}\bigr) \\ - \ \omega_s \nabla_{x_t} \, r\bigl(x_t \mid x^{<k}_{t}, \textrm{txt}\bigr),
\end{aligned}
\end{equation}
where we combine VJEPA surprise guidance with classifier-free guidance from both text and previous denoised chunks $x^{<k}_{t}$. In particular, we use previous denoised chunks as context for the VJEPA predictor, predict the next chunk, and calculate VJEPA's surprise reward.

\subsection{VLM-based Reward Model Details}

For VLM-based reward models, we use Qwen2.5-VL-7B-Instruct~\citep{bai2025qwen2} and Qwen3-VL-8B-Instruct~\citep{yang2025qwen3}, respectively. We use
a question template
\texttt{"Does the video show good physics dynamics and showcase a good alignment with the physical world? Please be a strict judge. If it breaks the laws of physics, please answer 0. Answer 0 for No or 1 for Yes. Reply only 0 or 1."}. Then, we extract the logit of token \texttt{"1"} and its variation \texttt{" 1"} as the reward signal. 

\section{Human Study Details}
\label{apx:human_details}
\begin{figure*}[t]
  \centering
    \includegraphics[width=0.88\linewidth]{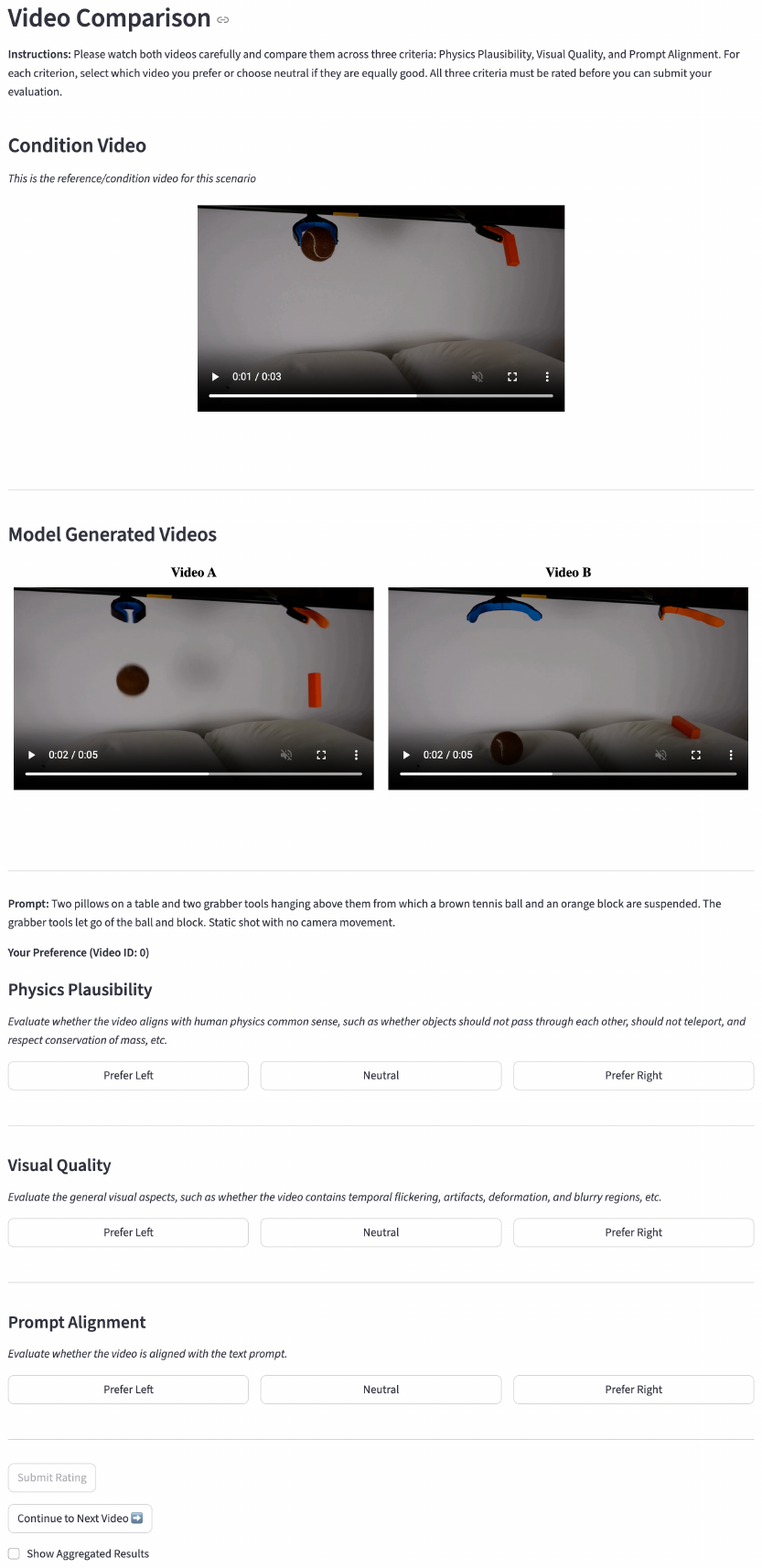}
    \vspace{-3mm}
\caption{\textbf{Human Study Interface.} Annotators view a side-by-side video comparison and indicate their preference on three criteria—Physics Plausibility, Visual Quality, and Prompt Alignment—choosing one of three preference options: Left, Right, or Neutral.}

  \label{fig:human_interface}
  \vspace{-3mm}
\end{figure*}

As shown in \Cref{fig:human_interface}, annotators are presented with a side-by-side comparison interface where they view two generated videos along with the original text prompt and conditioning frames describing the physical scenario. For each video pair, annotators provide judgments across three criteria: \textbf{(1) Physics Plausibility}, assessing whether the physical interactions and dynamics are realistic; \textbf{(2) Visual Quality}, evaluating the overall visual fidelity, clarity, and aesthetics of the generated video; and \textbf{(3) Prompt Alignment}, measuring how well the video content matches the given text description. 
For each criterion, the annotators select one of three options: preferring the video sample on the left, preferring the video sample on the right, or reporting a neutral preference when the difference is negligible or both videos are equally good/bad. 
To mitigate position bias, videos from each model are randomly assigned to the left or right positions. We collect evaluations from five annotators. Results are aggregated to compute win rates (percentage of comparisons where a model is preferred, excluding neutral judgments) and accuracy scores (computed as $\frac{\text{wins} + 0.5 \times \text{neutrals}}{\text{total}}$), providing a comprehensive assessment of relative model performance.\looseness-1

\section{Qualitative Examples}
\label{apx:visual_sample}

\begin{figure*}[t]
  \centering
    \includegraphics[width=0.9\linewidth]{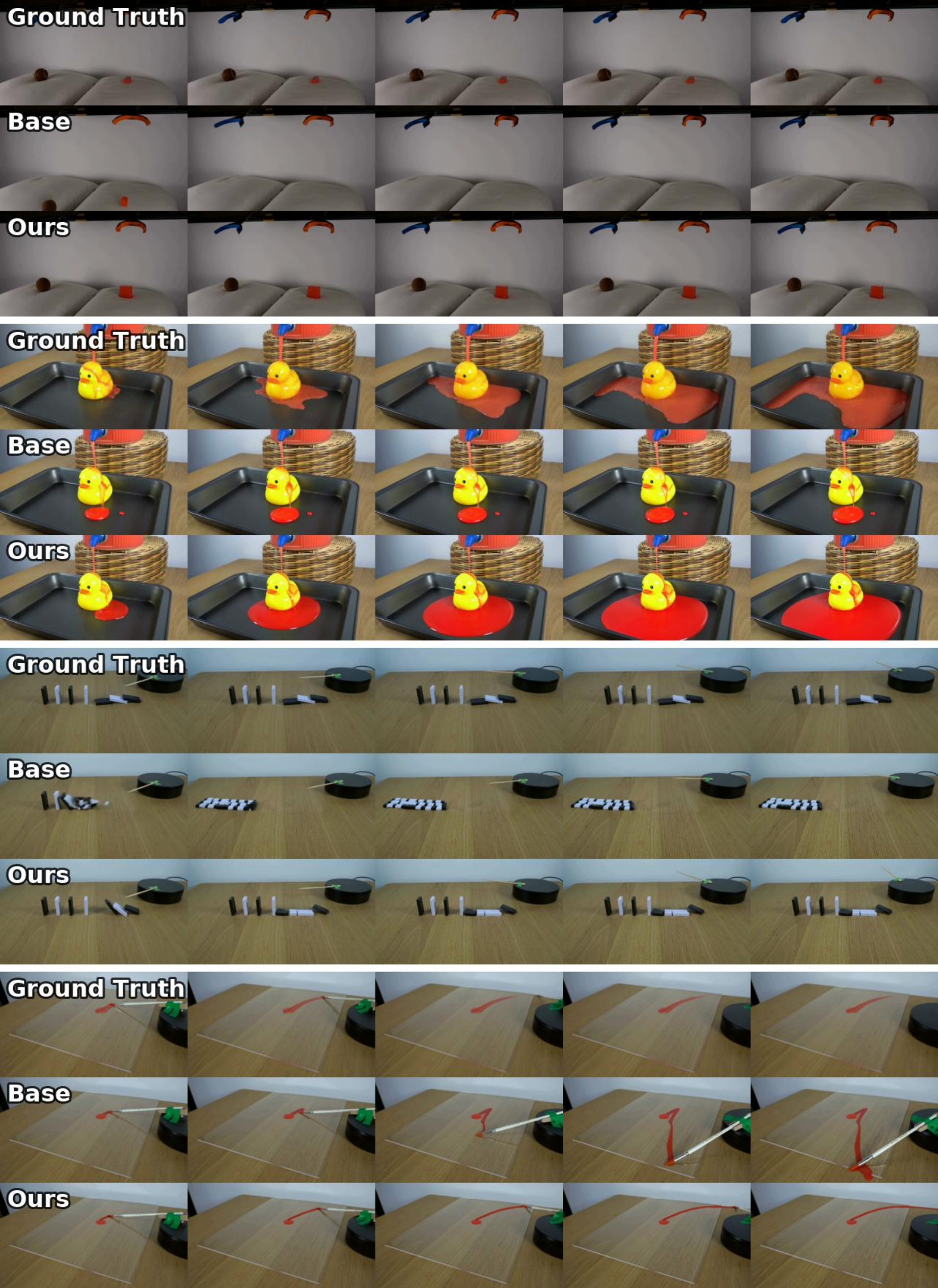}
  \caption{\textbf{Additional Qualitative Results on Physics-IQ.} 
  }
  \label{fig:more_visual1}
  \vspace{-3mm}
\end{figure*}

\begin{figure*}[t]
  \centering
    \includegraphics[width=0.9\linewidth]{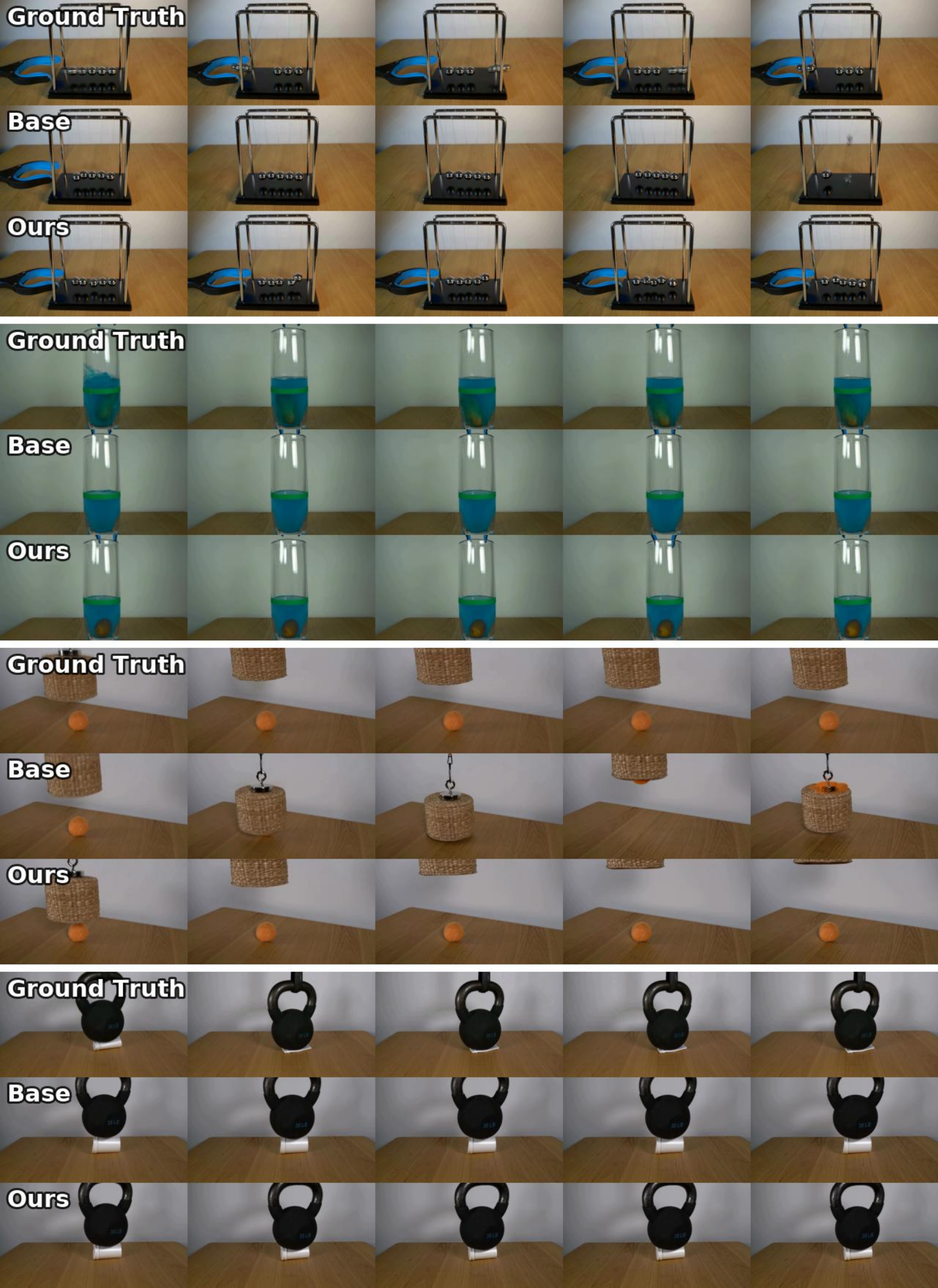}
  \caption{\textbf{Additional Qualitative Samples on Physics-IQ.} 
  }
  \label{fig:more_visual2}
  \vspace{-3mm}
\end{figure*}

\begin{figure*}[t]
  \centering
    \includegraphics[width=0.9\linewidth]{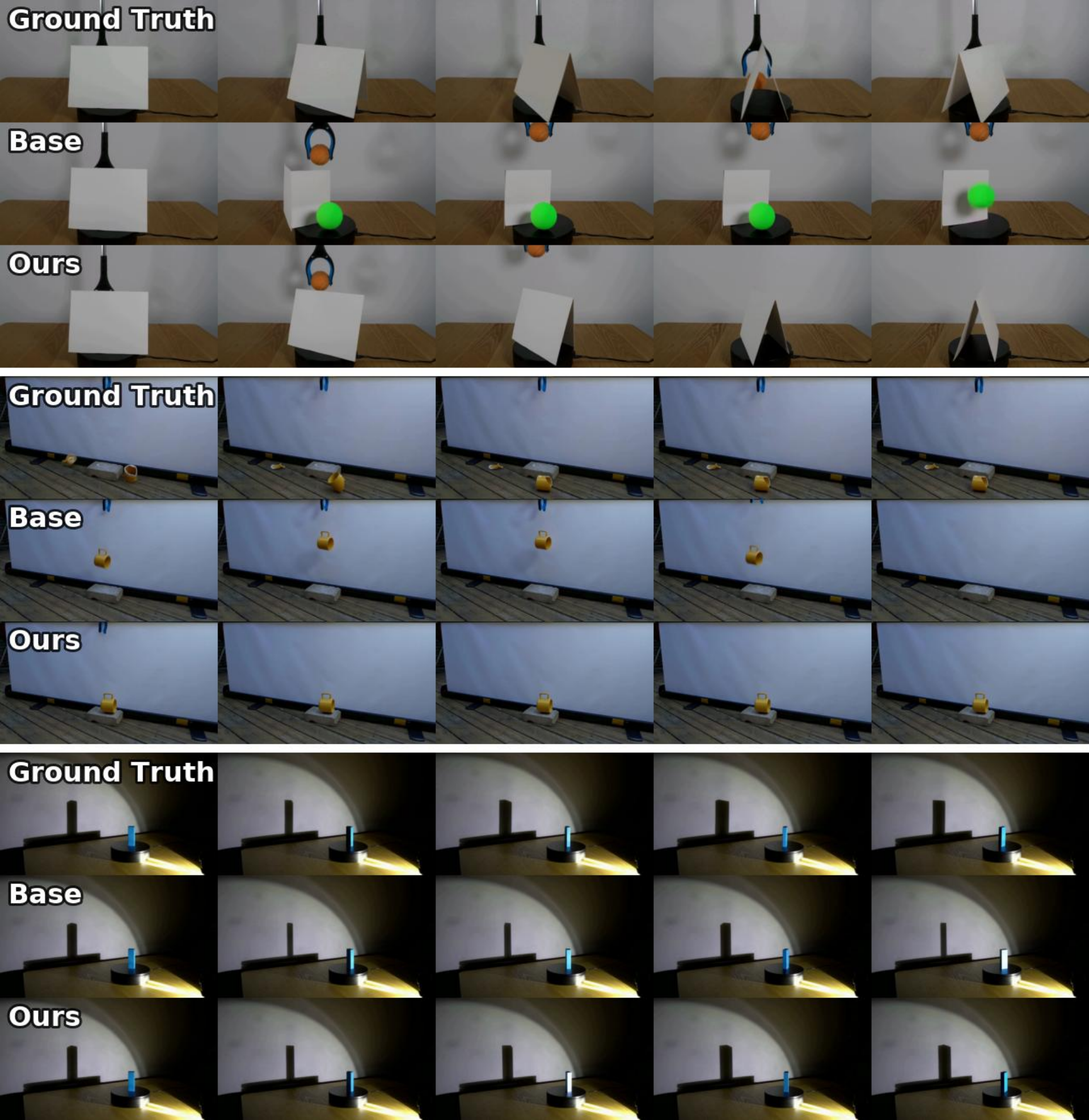}
  \caption{\textbf{Additional Qualitative Samples on Physics-IQ.} 
  }
  \label{fig:more_visual3}
  \vspace{-3mm}
\end{figure*}

\begin{figure*}[t]
  \centering
    \includegraphics[width=0.9\linewidth]{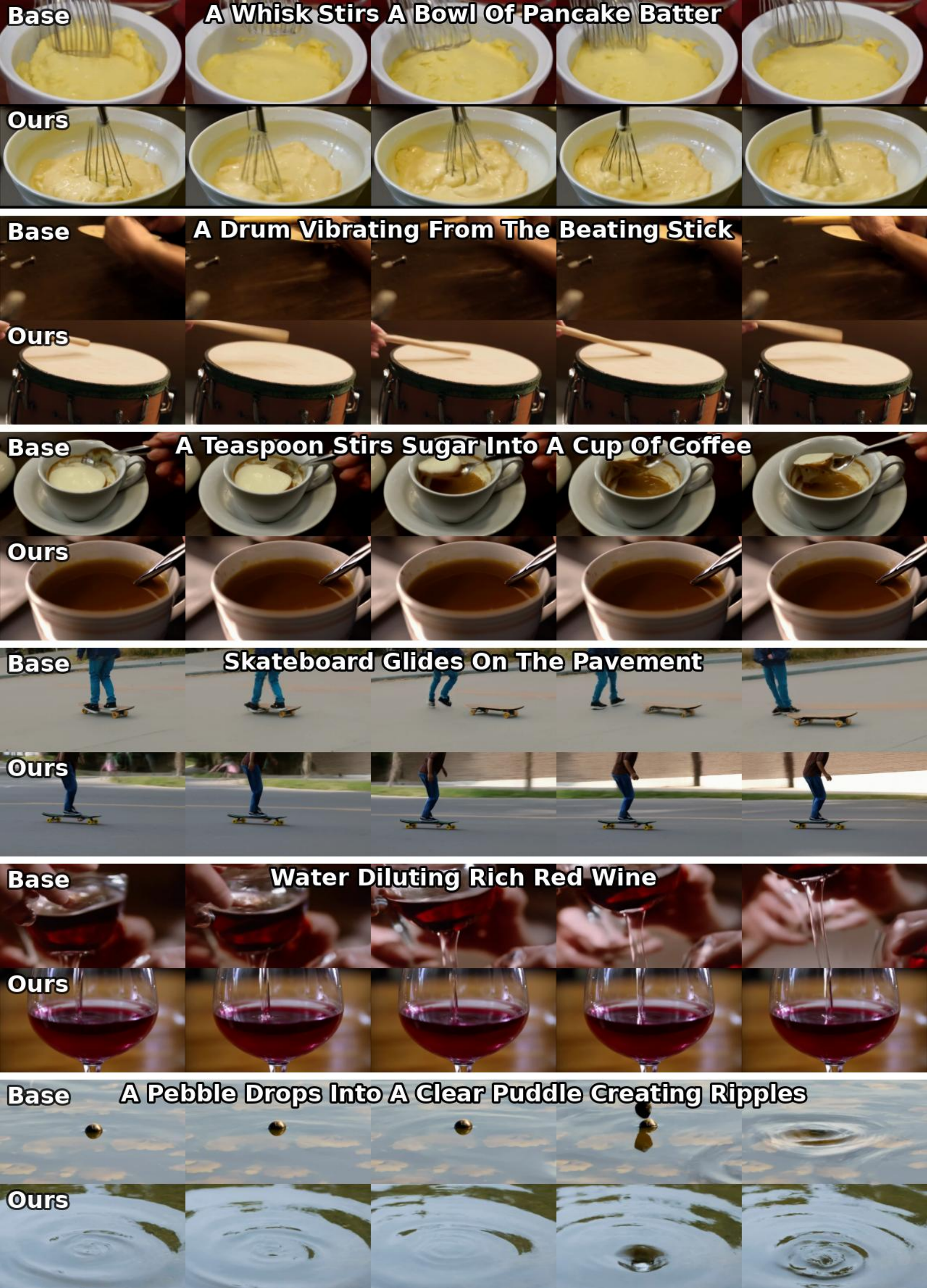}
  \caption{\textbf{Additional Qualitative Samples on VideoPhy.} 
  }
  \label{fig:more_visual4}
  \vspace{-3mm}
\end{figure*}

We present additional qualitative visual samples to demonstrate the effectiveness of \method. As shown in~\Cref{fig:more_visual1,fig:more_visual2,fig:more_visual3} for image- and multiframe-conditioned generation, and in~\Cref{fig:more_visual4} for text-conditioned generation, we observe that our generated videos exhibit improved physics plausibility across spatial continuity, rigid-body dynamics, fluid behavior, buoyancy, temporal continuity, gravity, conservation of mass, and optical effects. These samples indicate that the latent world model contains non-trivial physics understanding that enables more physically plausible video generation. \textit{For a better view of the dynamics, we recommend viewing the videos attached in the supplementary material zip file}.\looseness-1

\section{Failure Mode Analysis}
\label{apx:add_results}

\begin{figure*}[ht]
  \centering
    \includegraphics[width=\linewidth]{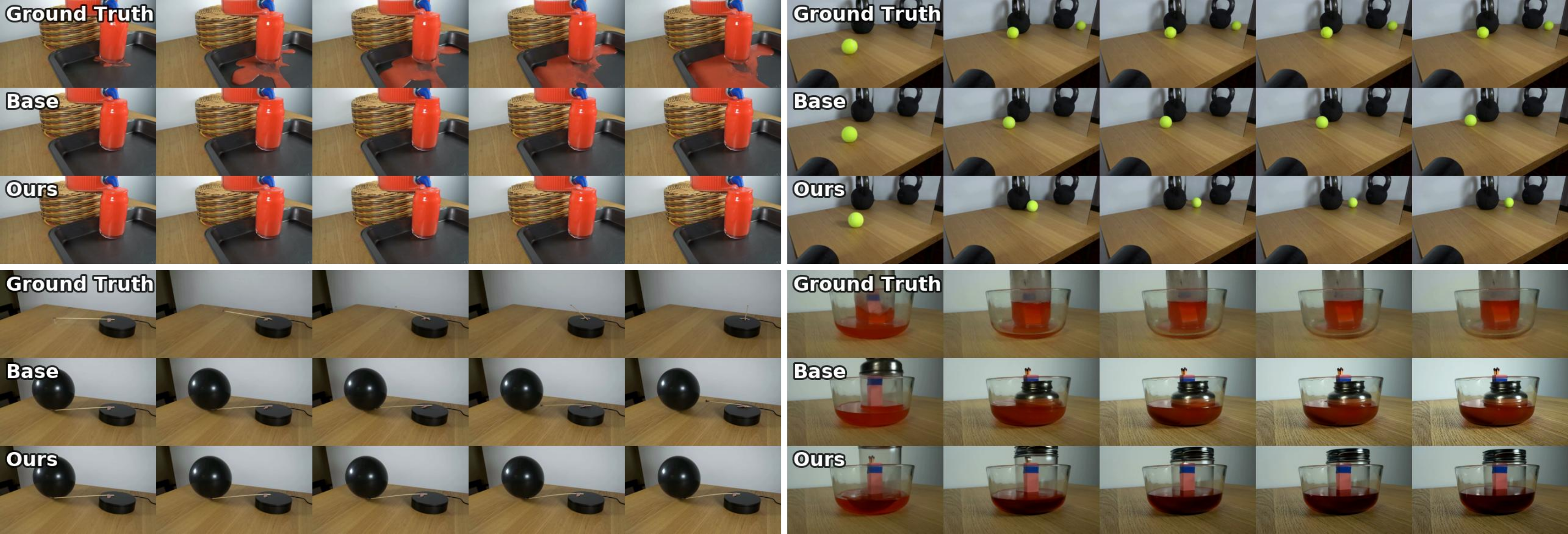}
\caption{\textbf{Failure Mode Analysis.} We observe some failure modes that persist even when leveraging VJEPA-2 for sampling. For example, the model often fails on abrupt physical events, such as fluid overflowing from a bottle (top left quadrant) or a lit match igniting a balloon and causing it to explode (bottom left quadrant). It also struggles with more complex phenomena that requires reasoning and understanding of material properties, including mirror reflections (top right) and siphon effects (bottom right), indicating that both the base model and the reward model still have room of improvement on physics understanding.} %
  \label{fig:failure}
\end{figure*}

One of our core assumptions is that the latent world model VJEPA-2 captures stronger intuitive physics than current video generators~\citep{garrido2025intuitive}. However, this learned prior remains a proxy for true physics dynamics. VJEPA surprise is not exclusively measuring physics plausibility and entangles other perceptual factors. 
Practically, as shown in \Cref{fig:failure}, we observe that sampling with VJEPA reward in some cases does not lead to substantial physics plausibility improvements. For example, the model fails to capture abrupt physical events, such as fluid overflowing from a bottle or a lit match igniting a balloon and causing it to explode, which require reasoning about sudden state changes. It sometimes struggles with more complex phenomena, including mirror reflections and siphon effects, which demand more complex reasoning and understanding of material properties. While VJEPA reward can correct some physics violations (\eg, conservation-of-mass errors in certain siphon scenarios), these failures indicate that there remains room to improve the physics understanding of latent world models in order to obtain more reliable reward signals and, consequently, better physics-aware video generation.\looseness-1

\end{document}